*Engineering Note*

# Engineering a Conformant Probabilistic Planner

**Nilufer Onder**                                          NILUFER@MTU.EDU
**Garrett C. Whelan**                                    GCWHELAN@MTU.EDU
**Li Li**                                                        LILI@MTU.EDU
*Department of Computer Science*
*Michigan Technological University*
*1400 Townsend Drive*
*Houghton, MI 49931*

## Abstract

We present a partial-order, conformant, probabilistic planner, Probapop which competed in the blind track of the Probabilistic Planning Competition in IPC-4. We explain how we adapt distance based heuristics for use with probabilistic domains. Probapop also incorporates heuristics based on probability of success. We explain the successes and difficulties encountered during the design and implementation of Probapop.

## 1. Introduction

Probapop is a *conformant probabilistic planner* that took part in the probabilistic track of the 4th International Planning Competition (IPC-4). It was the only conformant planner that competed. In the conformant probabilistic planning paradigm (Hyafil & Bacchus, 2003) the actions and the state initialization can be *probabilistic*, i.e., they can have several possible outcomes annotated by a probability of occurrence. In addition, the planning problem is *conformant*, i.e., the planner has to construct the best plan possible without assuming that the results of the actions performed can be observed. As an example of a conformant probabilistic planning problem, consider a student applying for graduate studies. Suppose that the application needs to include several forms prepared by the student and a single letter of recommendation written by a professor (one letter is sufficient but more than one letter is acceptable). Further assume that a typical professor in the student's department has 80% probability of sending a letter on time. In such a problem if the student asks one professor for a letter, the probability of having a complete application is 0.8. If the student observes that the professor has not sent a letter by the due date, there is no way to complete the application because it would be too late to ask another professor. Thus, observation actions are useless and the only way the student can increase the chances of getting a letter is to ask more than one professor to send in a letter. If 9 professors are asked, the probability of getting a letter is 0.999997 which is very close to 1. Obviously, asking too many people is costly, therefore the student has to weigh the benefits of increased probability against the costs of asking several people.

A conformant probabilistic planner's task is to find the best sequence of actions when the possible results of actions have predefined probabilities but cannot be observed. In that regard, conformant probabilistic planners can be classified as *non-observable Markov* decision processes (NOMDPs) (Boutilier, Dean, & Hanks, 1999). *Fully-observable MDPs (FOMDP)* are the other extreme of MDPs where the agent has complete and cost-free sensors that indicate the current state. Planners that adopt the FOMDP framework can generate policies that are functions from states to actions. NOMDP based planners can only generate unconditional sequences of actions based on a predictive model, because the environment cannot be observed (Boutilier et al., 1999). The middle ground is partially observable MDPs (POMDPs) and contingency plans where only some of the domain is observable and the execution of actions may depend on the results of observations (Kaelbling, Littman, & Cassandra, 1998; Majercik & Littman, 1999; Onder & Pollack, 1999; Hansen & Feng, 2000; Karlsson, 2001; Hoffmann & Brafman, 2005). There are also conformant planners which model imperfect actions that may have multiple possible results but do not model probability information (Ferraris & Giunchiglia, 2000; Bertoli, Cimatti, & Roveri, 2001; Brafman & Hoffmann, 2004).

Our work on Probapop is motivated by the incentive to have partial-order planning as a viable option for conformant probabilistic planning. The main reasons are threefold. First, partial-order planners have worked very well with parametric or lifted actions, which are useful in coding large domains. Second, due to its least commitment strategy in step ordering, partial-order planning (POP) produces plans that are highly parallelizable. Third, many planners that can handle rich temporal constraints have been based on the POP paradigm (Smith, Frank, & Jonsson, 2000). Given these advantages, our intuition in the design of Probapop was to bring together two paradigms that do not model states explicitly: POP planners do not represent states because they search in a space of plans, and blind planners cannot observe the state because no observation actions are available.

Our basic approach is to form base plans by using deterministic partial-order planning techniques, and then to estimate the best way to improve these plans. Recently, the Repop (Nguyen & Kambhampati, 2001) and Vhpop (Younes & Simmons, 2003) planners have demonstrated that the same heuristics that speed up non-partial-order planners can be used to scale up partial-order planning. We show that distance-based heuristics (McDermott, 1999; Bonet & Geffner, 2001) as implemented using "relaxed" plan graphs in partial-order planners such as Repop and Vhpop can be employed in probabilistic domains. These heuristics coupled with selective plan improvement heuristics and incremental planning techniques result in significant advantages. As a result, Probapop makes partial-order planning feasible in probabilistic domains. Our work on Probapop has been invaluable in understanding and identifying the key solutions to issues in probabilistic conformant planning.

## 2. Probapop and Partial-Order Planning

For partial-order probabilistic planning, we implemented the Buridan (Kushmerick, Hanks, & Weld, 1995) probabilistic planning algorithm on top of Vhpop (Younes & Simmons, 2003), a recent partial-order planner. A partially ordered plan $\pi$ is a 6-tuple, <STEPS, BIND, ORD, LINKS, OPEN, UNSAFE>, representing sets of actions, binding constraints, ordering constraints, causal links, open conditions, and unsafe links, respectively. A binding constraint is a constraint between action parameters and other action parameters or ground literals. An ordering constraint $S_i \prec S_j$ represents the fact that step $S_i$ precedes $S_j$. A *causal link* is a triple $< S_i, p, S_j >$, where $S_i$ is the *producer* step, $S_j$ is the *consumer* step and $p$ represents the condition supported by $S_i$ for $S_j$. An *open condition* is a pair $< p, S >$, where $p$ is a condition needed by step $S$. A causal link $< S_i, p, S_j >$ is *unsafe* if the plan contains a *threatening* step $S_k$ such that $S_k$ has $\neg p$ among its effects, and $S_k$ may intervene between $S_i$ and $S_j$. Open conditions and unsafe links are collectively referred to as *flaws*. A *planning problem* is a quadruple $< D, I, G, T >$, where D is a domain theory consisting of (probabilistic) operators, the initial state $I$ is a probability distribution over states, $G$ is a set of literals that must be true at the end of execution, and $T$ is a termination criterion such as a probability threshold or a time limit. The objective of the planner is to find the maximal probability plan that takes the agent from $I$ to $G$. If several plans have the same probability of success, then the one with the least number of steps or cost is preferred.

The Probapop algorithm shown in Figure 1 is based on the classical POP algorithm (Russell & Norvig, 2003; Younes & Simmons, 2003). It first constructs an initial plan by converting *initial* and *goal* into dummy initial and goal steps, and using those as the first and last steps of a plan with an empty body. It then refines the plans in the search queue until it meets the termination criterion. The termination criterion that were implemented include a time limit (e.g., stop after 5 minutes), a memory limit (e.g., stop after 256MB), a probability threshold (e.g., stop after finding a plan with 0.9 or higher probability), and lack of significant progress (e.g., stop if the probability of success cannot be increased more than $\epsilon$). It is possible to specify multiple termination criterion and use the earliest one that becomes true. When a termination criterion is met the plan with the highest probability is returned.

Plan refinement operations involve repairing flaws. An open condition can be closed by adding a new step from the domain theory, or reusing a step already in the plan. An unsafe link is handled by the *promotion*, *demotion*, or *separation* (when lifted actions are used) operations, or by *confrontation* (Penberthy & Weld, 1992). All of these techniques are part of the Vhpop implementation. Consider a step $S_k$ threatening a causal link $< S_i, p, S_j >$. Promotion involves adding an extra ordering constraint such that $S_k$ comes after $S_j$ ($S_j \prec S_k$ is added to ORD). Demotion involves adding an extra ordering constraint such that $S_k$ comes before $S_i$ ($S_k \prec S_i$ is added to ORD). Separation involves adding an extra inequality constraint to BIND such that $S_k$'s threatening effect can no longer unify with $\neg p$. Finally, when actions have multiple effects, confrontation can be used by making a commitment to non-threatening effects of $S_k$, i.e., those effects of $S_k$ that do not contain a proposition that unifies with $\neg p$. Note that in deterministic domains, an action can have multiple effects due to multiple secondary preconditions (*when* conditions). In probabilistic domains, probabilistic actions always have multiple effects.

The search is conducted using an A* algorithm guided by the *ranking function* which provides the $f$ value. As usual for a plan $\pi$, $f(\pi) = g(\pi) + h(\pi)$, where $g(\pi)$ is the cost of the plan, and $h(\pi)$ is the estimated cost of completing it. The ranking function is used at the MERGE step of the algorithm to order the plans in the search queue. In the competition Probapop used a distance based heuristic (*ADD*) as explained in the next section. For the flaw selection strategy in the SELECT-FLAW method, it used Vhpop's *static*, which gives priority to static open conditions, i.e., a condition whose value is not altered by any action in the domain theory. If the flaws of a plan do not contain any static open conditions threats are handled next; the lowest priority is given to the remaining open conditions. We **function** PROBAPOP (*D, initial, goal, T*)
**returns** a solution plan, or failure
    *plans* ← MAKE-MINIMAL-PLAN(*initial, goal*)
    *BestPlan* ← null
    **loop do**
      **if** a termination criterion is met **then return** BestPlan
      **if** *plans* is empty **then return** failure
      *plan* ← REMOVE-FRONT(*plans*)
      **if** SOLUTION?(*plan*) **then return** *plan*
      *plans* ← MERGE(*plans*, REFINE-PLAN(*plan*))
    **end**

**function** REFINE-PLAN (*plan*)
**returns** a set of plans (possibly null)
    **if** FLAWS(*plan*) is empty **then**
      **if** PROBSUCCESS (*plan*) > PROBSUCCESS (*BestPlan*)
        *BestPlan* ← *plan*
      *plan* ← REOPEN-CONDITIONS(*plan*)
    *flaw* ← SELECT-FLAW(*plan*)
    **if** *flaw* is an open condition **then choose**:
        **return** REUSE-STEP(*plan, flaw*)
        **return** ADD-NEW-STEP(*plan, flaw*)
    **if** *flaw* is a threat **then choose**:
        **return** DEMOTION(*plan, flaw*)
        **return** PROMOTION(*plan, flaw*)
        **return** SEPARATION(*plan, flaw*)
        **return** CONFRONTATION(*plan, flaw*)

Figure 1: The probabilistic POP algorithm.

comment on other heuristics and flaw selection techniques following the discussion of the competition results.

In the deterministic POP algorithm, a plan is considered to be *complete* when it has no flaws, i.e., OPEN = UNSAFE = $\emptyset$. In probabilistic domains, there is a possibility that complete plans that have insufficient probability of success (e.g., below $1 - \epsilon$) can be improved. Probapop improves such plans by conducting a search after reopening the conditions that can fail as explained in the next section. Probapop can be viewed as first searching for a plan that is complete in the deterministic sense, and then searching for a way to improve the plan. In our current implementation, we discard the search queue after finding the first plan and all the subsequent improvements are made on the first complete plan found. In the future, we plan to implement multiple search queues in order to be able to jump between different plans and their improvements. In Figure 2a, we show an initial plan that corresponds to the student application domain mentioned in the first section. The open conditions are sending the forms (forms-sent) and getting a letter of reference (letter-sent). Probapop uses Vhpop guided by the ranking and flaw selection heuristics to produce a complete plan with 80% probability of success shown in Figure 2b. A straight line shows a causal link between two actions and a zigzag line refers to a causal link from a plan fragment that has been omitted for clarity of exposition. Probapop reopens the condition "letter-sent" (Figure 3a) and resumes its search using the same heuristics to come up with an improved plan that involves asking two professors as shown in Figure 3b. Assuming that `ASK-PROFx` is the only action that has probabilistic effects, the probability of success is 0.8 for the first complete plan and $0.8 + 0.2 \times 0.8$ for the second complete plan. Several such iterations of reopen and search leads Probapop to find a plan with a probability of 0.999997. Such a plan cannot be improved further with single precision arithmetic.

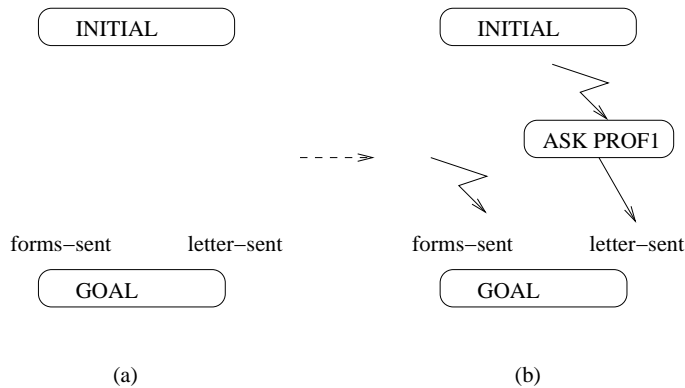

Figure 2: Starting with an empty plan and finding a first plan.

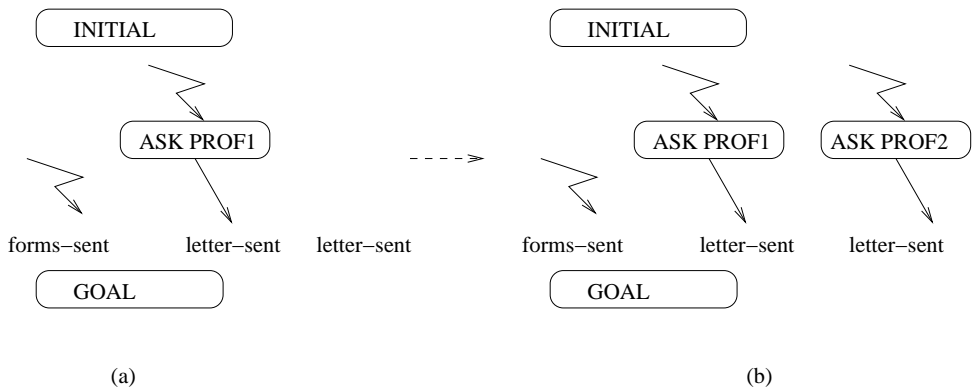

Figure 3: Starting with a complete plan and finding an improved plan.

## 3. Distance Based Ranking in Probapop

The Vhpop deterministic partial order-planner described by Younes and Simmons (2003) supports distance based heuristics to provide an estimate of the total number of new actions needed to close an open condition. Before starting to search, the planner builds a planning graph (Blum & Furst, 1997), which has the literals in the initial state in its first level, and continues to expand the graph until it reaches a level where all the goal literals are present. The planning graph is different than Graphplan's planning graph in the sense that it is *relaxed*, i.e., delete lists are ignored and thus mutex relationships are not computed (Bonet & Geffner, 2001).

In order to be able to generate a relaxed planning graph when multiple probabilistic effects are present, one would need to split into as many plan graphs as there are leaves in a probabilistic action. To avoid this potential blow up, we split each action in the domain theory into as many deterministic actions as the number of nonempty effect lists. Each split action represents a different way the original action would work. In Figure 4, we show an action A1, which has two probabilistic effects $a$ and $b$ when $P$ and $Q$ are true, one effect $c$ when $P$ is true and $Q$ is false, and no effect otherwise. Each split action corresponds to one set of non-empty effects. In Probapop, while the plan graph uses split actions, the plans constructed always contain the full original action so that the planner can correctly assess the probability of success. By using the split actions, we can compute a good estimate of the number of actions needed to complete a plan for use with distance based heuristics.

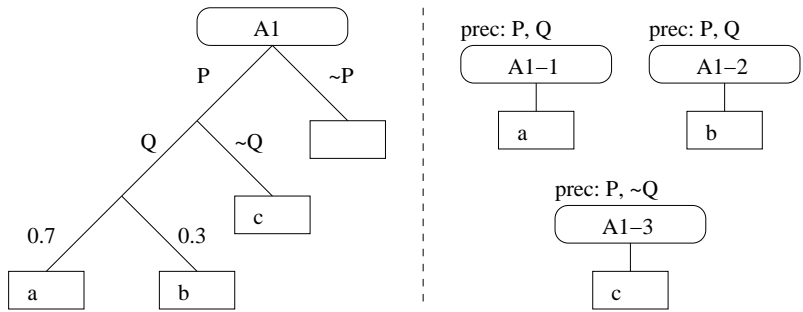

Figure 4: Probabilistic action A1 is split into deterministic actions A1-1, A1-2, and A1-3.

An important distinction between deterministic partial-order planning and probabilistic partial-order planning is multiple support for plan literals. In the deterministic case, an open condition is permanently removed from the list of flaws once it is resolved. In the probabilistic case, it can be *reopened* so that the planner can search for additional steps that increase the probability of the literal. The Buridan system implements this technique by reopening all the previously closed conditions of a complete plan and resuming the search to find another complete plan. Our implementation employs *selective reopening* (SR) where only those conditions that are not guaranteed to be achieved are reopened. In other words, literals supported with a probability of 1 are not reopened. Note that while checking the probability of literals is costly for probabilistic plans, we save most of the cost by performing the check during mandatory assessment of complete plans. Obviously, avoiding redundant searches is an advantage for the planner. In our current implementation we reopen all the supported literals that have a probability less than 1. We leave the selection from among this new set of preconditions to the flaw selection heuristic. Our implementation does not contain any probability based heuristics.

It is important to note that neither the split actions nor the selective reopening technique change the base soundness and completeness properties of the Buridan algorithm. The split actions are only used in the relaxed plan graph, and the reopening technique does not block any alternatives from being sought as they would already be covered by a plan in the search queue.

## 4. Probapop in IPC-4

Probapop was among the 7 domain-independent planners that competed in the probabilistic track of IPC-4. By domain-independent we mean a planner that uses only the PPDDL description of a domain to solve a planning problem and does not employ any previously coded control information. In Table 1 we show a brief description of these planners (Edelkamp, Hoffman, Littman, & Younes, 2004; Younes, Littman, Weissman, & Asmuth, 2005; Bonet & Geffner, 2005; Fern, Yoon, & Givan, 2006; Thiebaux, Gretton, Slaney, Price, & Kabanza, 2006). The competition was conducted as follows: Each planner was given a set of 24 problems written in probabilistic PDDL (PPDDL) and was allotted 5 minutes to solve the problem. After this, the server simulated a possible way of executing the plan by sending a sequence of states starting with the initial state and the planners responded to each state with an action based on the solution they found. 30 simulations were conducted for each problem. For *goal-based* problems success was measured by whether the goal was reached at the end of the simulation. For *reward-based* problems the total reward was calculated. The set of 24 problems included both of these types.

The competition included various domains as listed below:

- Blocksworld: Includes the pick up and put down actions where each action can fail. 6 problems with 5, 8, 11, 15, 18 and 21 blocks were given. The goal was to build one or more towers of blocks.

- Colored Blocksworld: The actions are the same as the Blocksworld domain. Each block can be one of three colors. The goal towers were specified using existential quantifiers, e.g., there is a green block on the table, there is a red block on a green block.

- Exploding Blocksworld: It is similar to the Blocksworld domain but the first put-down action can permanently destroy the bottom object (block or table). Replanning or repetition based approaches fail easily due to the irreversible nature of the explosion.

- Boxworld: It is a box transportation problem with load, unload, drive and fly actions. The drive action can fail taking the truck to a wrong city.

- Fileworld: The objective includes actions to put the papers into files of matching type. The type of a paper can be found out by using an observation action that has probabilistics outcomes.

- Tireworld: The actions include moving between several cities and the tire can go flat during a trip.

- Towers of Hanoise: It is a variation of the Towers of Hanoi problem where discs can be moved in singles or doubles and discs may slip during a move.

- Zeno travel: It is a travel domain that includes actions related to flying. Some actions such as boarding and flying can fail.

It should be noted that all of the competition domains were designed for full observability and needed to be changed to incorporate a blind planner. For instance, the PICKUP action in the Blocksworld domain has a precondition that requires that the block to be picked up is not being held by the arm. The action has two probabilistic effects, one resulting in the block being held, and the other being not held. Because the planner assumes no observability, a plan involving a PICKUP action cannot be improved because an action cannot be executed unless its preconditions hold. Thus, the Probapop planner cannot insert a second PICKUP action to cover the case in which the first one fails. With the help of the competition organizers, we implemented a workaround such that the actions that are executed when their conditions do not hold have no effect rather than causing an error.

| Planner (code) | Description |
| --- | --- |
| UMass (C) | Symbolic heuristic search based on symbolic AO* with loops (LAO*) and symbolic real-time dynamic programming (RTDP) |
| NMRDPP (G1) | Solving decision problems with non-Markovian (and hence Markovian) rewards |
| Classy (J2) | Approximate policy iteration with inductive machine learning using random-walk problems |
| FF-rePlan (J3) | Deterministic replanner using Fast Forward |
| mGPT (P) | Labeled real-time dynamic programming (LRTDP) with lower bounds extracted from the deterministic relaxations of the MDP |
| Probapop (Q) | POP-style plan-space A* search with distance based heuristics and failure analysis |
| CERT (R) | Heuristic state space search with structured policy iteration algorithm, factored MDPs, and reachability analysis |

Table 1: Domain-independent planners listed in order of competition code.

Probapop (competition name Q) attempted 4 of the 24 problems. The two planners that attempted most of the problems were Classy (J2) and FF-rePlan (J3). The other planners attempted between 3 to 10 problems as listed in Table 2. Probapop attempted a small number of problems due to three reasons. First, when we started building Probapop, Vhpop's version was 2.0. The performance of Vhpop was significantly improved with better memory handling techniques in version 2.2 but we did not have time before the competition to convert our implementation to the newer version. Second, the competition Blocksworld domains included universally quantified preconditions which were not supported in Vhpop. Our implementation of the preconditions including the FORALL keyword was not efficient. Third, our implementation disables the feature of Vhpop which allows the use of multiple search queues with different heuristics. This prohibited us from constructing several search queues each with a different heuristic and using the one that finishes the earliest. We therefore had to pick a single heuristic to run the competition problems. As a result, we picked *ADD* as the ranking metric and *static* as the flaw selection technique and ran all the problems with this combination.

After the competition results were announced, we observed that there were three domain independent planners, namely Classy (J2), FF-rePlan (J3), and mGPT (P), that were able to solve the largest Blocksworld problems whereas Probapop was only able to solve the 5-blocks problem (the competition included domains with 5, 8, 11, 15, 18, and 21 blocks). Therefore, we looked for ways of improving the performance of Probapop on these problems. We first reimplemented Probapop on Vhpop's newer version 2.2. Second, we brought the language of the competition Blocksworld domain closer to STRIPS. In particular, we removed the FORALL preconditions and WHEN conditions. For example, we replaced the PPDDL PICK-UP action shown in Figure 5 with the two actions shown in Figure 6. However, the version upgrade and the language simplification were not sufficient to enable Probapop to solve the 8-blocks problem. As explained before, Probapop's strategy is to first find a "base plan" and then to improve this plan at possible failure points, therefore finding the base plan is crucial. We next looked for other heuristics and flaw selection strategies that can make the Blocksworld problems solvable. We begin discussing these by explaining Vhpop's ADD heuristic in more detail.

| Planner | # of problems | bw-nc-r-5 | tire-nr | tire-r | zeno |
|---|---|---|---|---|---|
| Umass (C) | 4 | 30 | 30 | 30 | 30 |
| NMRDPP (G1) | 7 | 30 | 9 | 30 | 30 |
| Classy (J2) | 18 | 30 | – | – | – |
| FF-rePlan (J3) | 24 | 30 | 7 | 30 | 0 |
| mGPT (P) | 10 | 30 | 16 | 30 | 30 |
| **Probapop (Q)** | **4** | **11** | **7** | **6** | **1** |
| CERT (R) | 3 | 30 | 9 | 0 | 27 |

Table 2: The number of successes in 30 trials obtained by the planners that do not use domain knowledge. Only the problems attempted by Probapop (**Q**) are listed. A dash means that the planner did not attempt that problem. Bw-nc-r-5 is the Blocksworld problem with 5 blocks. Tire-nr and tire-r are the goal and reward based problems from the Tireworld domain. Zeno is a problem using the Zeno travel domain problem.

```
(:action pick-up-block-from
    :parameters (?top - block ?bottom)
    :effect (when (and (not (= ?top ?bottom)) (on-top-of ?top ?bottom)
                        (forall (?b - block) (not (holding ?b)))
                        (forall (?b - block) (not (on-top-of ?b ?top))))
                   (and (decrease (reward) 1)
                   (probabilistic 0.75 (and (holding ?top) (not (on-top-of ?top ?bottom)))
                               0.25 (when (not (= ?bottom table))
                                      (and (not (on-top-of ?top ?bottom)) (on-top-of ?top table)))))))))
```

Figure 5: PPDDL's PICK-UP action

The ADD heuristic achieves good performance by computing the sum of the step costs of the open conditions from the relaxed planning graph, i.e., the heuristic cost of a plan is computed as $h(\pi) = h_{add}(OPEN(\pi))$. The cost of achieving a literal q is the level of the first action that achieves q: $h_{add}(q) = min_{a \in GA(q)} h_{add}(a)$ if $GA(q) \neq \emptyset$, where $GA(q)$

```
(:action pick-up
    :parameters (?x)
    :precondition (and (clear ?x) (ontable ?x) (handempty))
    :effect
        (probabilistic 0.75
          (and (not (ontable ?x)) (not (clear ?x)) (not (handempty)) (holding ?x))))

(:action unstack
    :parameters (?x ?y)
    :precondition (and (on-top-of ?x ?y) (clear ?x) (handempty))
    :effect
        (probabilistic 0.75
          (and (holding ?x) (clear ?y) (not (clear ?x)) (not (handempty)) (not (on-top-of ?x ?y)))))
```

Figure 6: Simplified form of PPDDL's PICK-UP action.

is an action that has an effect $q$. Note that $h_{add}(q)$ is 0 if $q$ holds initially, and is $\infty$ if $q$ never holds. The *level* of an action is the first level its preconditions become true: $h_{add}(a) = 1 + h_{add}(PREC(a))$. The *ADDR* heuristic is a modification of the ADD heuristic that takes action reuse into account, thus in addition to the conditions described above, the heuristic cost of a literal $q$ is 0 if the plan already contains an action that can achieve $q$.

We observed that ADDR is more effective than ADD for the Blocksworld domain and tested a variety of flaw selection strategies implemented in Vhpop together with ADDR. We show the flaw selection strategies we tried in Table 3. We adopt the notation given by Pollack et al. (1997) and revised by Younes and Simmons (2003). In this notation, each strategy is an ordered list of selection criteria where LR refers to "least refinements first", MC$_{add}$ refers to "most cost computed using ADD", and MW$_{add}$ refers to "most work using ADD". Open conditions are divided into three categories for use by some heuristics. A *static open condition* is an open condition whose literal can only be provided by the initial state, i.e., no action has this literal as an effect. A *local open condition* refers to the open conditions of the most recently added action and is used to maintain focus on the achievement of a single goal. An *unsafe open condition* refers to an open condition whose causal link would be threatened.

There are five main strategies which prioritize flaws differently. The *ucpop strategy* gives priority to threats, the *static strategy* gives priority to static open conditions, the *lcfr strategy* handles flaws in order of least expected cost, the *mc strategy* orders open conditions with respect to cost extracted from the relaxed planning graph, and the *mw strategy* orders open conditions with respect to expected work extracted from the relaxed planning graph. A strategy with a "loc" annotation gives priority to local open conditions among the open conditions, a strategy with a "conf" annotation gives priority to unsafe open conditions among the open conditions. We refer the reader to the paper by Younes and Simmons (2003) for a thorough description of these heuristics as well as experimental results with other domains.

We depict the results of our experiments with the Blocksworld problems in the first and third lines of Table 4 (the second and fourth lines in Tables 4 and 5 will be explained later).

| Strategy | Description |
| --- | --- |
| ucpop | {n,s} LIFO / {o} LIFO |
| static | {t} LIFO / {n,s} LIFO / {o} LIFO |
| lcfr | {n,s,o} LR |
| lcfr-loc | {n,s,l} LR |
| lcfr-conf | {n,s,u} LR / {o} LR |
| lcfr-loc-conf | {n,s,u} LR / {l} LR |
| mc | {n,s} LR / {o} $MC_{add}$ |
| mc-dsep | {n} LR / {o} $MC_{add}$ / {s} LR |
| mc-loc | {n,s} LR / {l} $MC_{add}$ |
| mc-dsep | {n} LR / {l} $MC_{add}$ / {s} LR |
| mw | {n,s} LR / {o} $MW_{add}$ |
| mw-dsep | {n} LR / {o} $MW_{add}$ / {s} LR |
| mw-loc | {n,s} LR / {l} $MW_{add}$ |
| mw-loc-dsep | {n} LR / {l} $MW_{add}$ / {s} LR |

Table 3: The description of a variety of flaw selection strategies in Vhpop. "n" is a non-separable threat, "s" is a separable threat, "o" is an open condition, "t" is a static open condition, "l" is a local open condition, and "u" is an unsafe open condition.

It can be seen that only lcfr and mc strategies work for the problem with 8 blocks. The larger problems were not solvable. Because the actions are lifted, we tried to make the search space smaller by delaying separable threats. Peot and Smith (1993) explain that delaying the separable threats may result in a decreased branching factor because there may be many ways to add inequality constraints for separation. The delay might also help because the threat can disappear as more variables are bound. We modified the best working strategies, namely variants of mc and mw, and implemented the delay of separable threats (in Table 3 these are shown with the *dsep* suffix.) We show the planning times for the experiments with and without dsep in Table 5 (we repeat the columns from Table 4 for comparison). The results show that time improvement can be seen for the 5-blocks problem. The problems with 8 blocks show an increase in time because each threat must be checked to see if it is separable. Delaying threats made the 8-blocks problem solvable using the mc-loc, mw, and mw-loc strategies. However, larger problems were not solvable by any strategy.

The results of our experiments with various heuristics and strategies show that the search time increases dramatically by going from 5 to 8 blocks and larger problems are not solvable. We were not able to find a heuristic combination to solve the larger problems. We noticed that the competition Blocksworld problems list the goal towers from top to bottom and the planner spends a lot of time with dead end plans when the original goal order is preserved. If a tower is built from top to bottom, the initial goals almost always have to be undone to achieve the later goals. We also concluded that such an interaction cannot be detected with the heuristics we used because they are designed to consider subgoals in isolation. Koehler and Hoffmann (2000) describe a polynomial time algorithm that can order goals to minimize the above type of undoing. The algorithm operates on ground action descriptions which can be generated from action schemas and was implemented in the FF planning system (Hoffman & Nebel, 2001). We used this algorithm to order the top-level goals and repeated all the experiments with this ordering which essentially builds towers from the bottom to the top. The results for ordered goals are shown in lines 2 and 4 of tables 4 and 5. Ordering the goals had mixed results. For example, for the 8 blocks problem, it made the lcfr heuristic not usable but the mw heuristic usable. However, the lowest time increased from 13K to 41K milliseconds and the larger problems were still not solvable.

| | ucpop | static | lcfr | lcfr-loc | lcfr-conf | lcfr-loc-conf | mc | mc-loc | mw | mw-loc | mw-loc-conf |
|---|---|---|---|---|---|---|---|---|---|---|---|
| 5 | 80 | 70 | 0 | 60 | 570 | 90 | 10 | 50 | 0 | 20 | 220 |
| 5o | 0 | 0 | 10 | 50 | 10 | 770 | 0 | 40 | 0 | 30 | 120 |
| 8 | – | – | 55K | – | – | – | 13K | – | – | – | – |
| 8o | – | – | – | – | – | – | 42K | – | 41K | – | – |

Table 4: Time (msec) required to find the base plan for Blocksworld problems with 5 and 8 blocks.

| | mc | mc-dsep | mc-loc | mc-loc-dsep | mw | mw-dsep | mw-loc | mw-loc-dsep |
|---|---|---|---|---|---|---|---|---|
| 5 | 10 | 0 | 40 | 20 | 0 | 0 | 20 | 30 |
| 5o | 0 | 0 | 50 | 30 | 0 | 0 | 30 | 20 |
| 8 | 13K | 73K | – | 22K | – | 73K | – | 22K |
| 8o | 42K | 104K | – | – | 41K | 103K | – | – |

Table 5: Time (msec) required to find the base plan by delaying separable threats for Blocksworld problems with 5 and 8 blocks.

Our final strategy was to combine the planning approach used by the FF planner with POP-style search. In particular, we ordered the top-level goals using FF's ordering algorithm and ran Vhpop $n$ times for problems with $n$ top level goals. The first problem had only the first goal and when Vhpop returned a plan, the steps were simulated to find the resulting state. The second problem had this resulting state as the initial state and goals 1 and 2 so that goal 1 would be preserved or redone and goal 2 would be achieved. When we used this strategy with the default heuristics of Vhpop to solve the problem with 21 blocks, the total time was 70 milliseconds with most phases taking 0 milliseconds. Koehler and Hoffmann (2000) explain that this approach works well for *invertible planning problems*, i.e., problems such as the Blocksworld where actions are reversible. In our case, the tradeoff is the possibility of less optimal plans because a plan for the $i$th goal is set while working on the $i + 1$st goal. The second tradeoff is getting several partially-ordered plans with breakpoints between problems rather than a single maximally parallel plan. We believe it is worthwhile to work on an algorithm that combines the individual plans to preserve the least commitment on ordering. Possible strategies are to causally link action preconditions to latest producers or use the approach of Edelkamp (2004) and parallelize sequential plans using critical path analysis.

## 5. Conclusion and Future Work

We presented the design and implementation of Probapop, a partial-order, probabilistic, conformant planner. We described the distance-based and condition-probability based heuristics that we used. We discussed the advantages and disadvantages of using an incremental algorithm where goals are first ordered and submitted one by one. Our short term plans involve implementing multiple search queues for different base plans and reincorporating the ADL constructs in PPDDL. Our future work involves three threads. In one, we are looking at improving the performance of Probapop by adding probability information to the planning graph so that the probability of open conditions can be optimistically estimated. We are also considering the addition of domain specific information (Kuter & Nau, 2005) to probabilistic domains. In the second thread, we are exploring the middle ground between no observability and full observability by considering POMDP-like problems in a partial-order setting. Finally, we would like to incorporate hill climbing techniques into our probabilistic framework. The current Probapop 2.0 software is available through www.cs.mtu.edu/~nilufer.

## Acknowledgments

This work has been supported by a Research Excellence Fund grant to Nilufer Onder from Michigan Technological University. We thank JAIR IPC-4 special track editor David E. Smith, and the anonymous reviewers for their very helpful comments.